# Weighted Unsupervised Domain Adaptation Considering Geometry Features and Engineering Performance of 3D Design Data

Seungyeon Shin[1], Namwoo Kang[1,2,*]

[1]Cho Chun Shik Graduate School of Mobility, KAIST, 34051, Daejeon, South Korea

[2]Narnia Labs, 34051, Daejeon, South Korea

[*]Corresponding author: nwkang@kaist.ac.kr

## Abstract

The product design process in manufacturing involves iterative design modeling and analysis to achieve the target engineering performance, but such an iterative process is time consuming and computationally expensive. Recently, deep learning-based engineering performance prediction models have been proposed to accelerate design optimization. However, they only guarantee predictions on training data and may be inaccurate when applied to new domain data. In particular, 3D design data have complex features, which means domains with various distributions exist. Thus, the utilization of deep learning has limitations due to the heavy data collection and training burdens. We propose a bi-weighted unsupervised domain adaptation approach that considers the geometry features and engineering performance of 3D design data. It is specialized for deep learning-based engineering performance predictions. Domain-invariant features can be extracted through an adversarial training strategy by using hypothesis discrepancy, and a multi-output regression task can be performed with the extracted features to predict the engineering performance. In particular, we present a source instance weighting method suitable for 3D design data to avoid negative transfers. The developed bi-weighting strategy based on the geometry features and engineering performance of engineering structures is incorporated into the training process. The proposed model is tested on a wheel impact analysis problem to predict the magnitude of the maximum von Mises stress and the corresponding location of 3D road wheels. This mechanism can reduce the target risk for unlabeled target domains on the basis of weighted multi-source domain knowledge and can efficiently replace conventional finite element analysis.

# 1. Introduction

At the conceptual design stage of industrial product development, engineering performance analysis and design modifications are performed iteratively until the design meets the target engineering performance. However, evaluating various designs at the conceptual design stage is challenging because computer simulations, such as finite element analysis (FEA), are time consuming and computationally expensive.

Given that deep learning can be used to approximate complex functions, recent studies have been conducted to efficiently replace computer simulations, such as FEA and computational fluid dynamics (CFD), with deep learning. Deep learning-based engineering performance prediction models are necessary for industrial applications because they can be used at the early stages of product development to examine various design alternatives in real time. This examination can effectively reduce the time and cost of product development. Inferences can be made directly on test data by training a deep learning model on existing simulation data.

However, these methods were developed under the assumption that future test data have the same distribution as training data. When the trained model is tested with new industrial data, its accuracy often decreases because the distribution of the training data and that of the new data to be tested are similar but somehow different, leading to poor predictions. Various studies have also observed this problem in the fields of computer vision and natural language processing.

The idea of domain adaptation was introduced to transfer knowledge obtained from the training data (source domain) to the test data (target domain). Similar to the idea of the generative adversarial network (GAN) (Goodfellow et al., 2020), an adversarial training strategy is used to align the source and target domains to a common latent space and minimize the distance between the domains through deep neural networks.

Domain adaptation can reduce the burden of recollecting and annotating a sufficient amount of data and training a new model every time similar but slightly different data emerge. It compensates for the limitations of deep learning models in data collection, and it can have a huge advantage, especially in industrial problems involving 3D data.

Industrial products with similar but diverse designs can be produced due to the complex features of a structure, which means that the data distribution may differ every time design variations are made. When a deep learning-based stress prediction model is built to replace FEA, the model needs to be trained on a large amount of FEA analysis results, so the front-loading time of data collection for a new target domain is inefficient. Therefore, constructing a deep learning model that can predict engineering performance regardless of the domain is advantageous, and establishing a domain adaptation method specialized for engineering analysis problems is useful.

However, the majority of existing unsupervised domain adaptation (UDA) studies have proposed methods for classification tasks, which have limited applicability in engineering performance predictions. In addition, although domain adaptation studies have been conducted on regression tasks, a domain adaptation method that considers the geometry features and engineering performance of 3D design data and can be directly applied to engineering analysis problems is still needed.

We aim to develop a deep learning-based engineering performance prediction model based on bi-weighted unsupervised domain adaptation (BW-UDA) that considers the geometry features and engineering performance of 3D design data to replace conventional 3D FEA. Generally, a regression task is performed on 3D design data to predict the engineering performance that must be inspected. The proposed model allows accurate predictions even for the unlabeled target domain, which is similar to the existing source domain but with a slightly different data distribution. The developed BW-UDA is trained with a source instance weighting strategy on the basis of the relation of the geometry features and engineering performance of the target and multi-source domain instances.

In particular, we experimentally apply the proposed method on a 3D wheel impact analysis problem. A deep learning-based prediction model is constructed to predict the magnitude of the maximum von Mises stress and the corresponding location (x, y, and z coordinates), and a 3D wheel voxel and an impact barrier mass are used as the input. We employ domain adaptation to leverage the knowledge of the labeled source domain wheel dataset and make accurate predictions for the unlabeled target domain (i.e., wheel dataset that has new rim cross sections). Given that each source sample has a different relation to the target domain, we apply appropriate weights to each source instance on the basis of the features of the 3D design data to avoid negative transfers. For this purpose, each source instance is weighted in consideration of two factors. One is based on the similarity of geometry features, and the other is based on the similarity of engineering performance features with the target domain. Training is performed with an adversarial approach by applying the combined weights to the source instances to reduce the target risk of the unlabeled target domain. With the proposed method, a generalized deep learning model can be obtained to predict wheel impact performance even for 3D wheels with unseen rim cross sections.

The contribution of this study is as follows:
1. This study is the first to apply domain adaptation to deep learning-based engineering performance prediction and use it to replace conventional 3D FEA. In particular, we show that UDA can efficiently replace wheel impact analysis, which requires time-consuming data collection. The proposed method can be extended to other engineering analysis problems.

2. The proposed bi-weighting strategy is specialized for 3D design data. No previously established UDA method considers geometry features and engineering performance simultaneously. We propose a BW-UDA method that reflects the similarity between target domain and source domain instances for both factors and show its considerable performance improvement relative to existing methods.

3. We transfer the knowledge of the multi-source domain to the target domain. Given the limited amount of industrial data for various domains, a method that utilizes multi-source domains can be established to solve the problem of insufficient data. In particular, a bi-weighting strategy allows the use of multi-source domain data even with a compact network.

The rest of this paper is organized as follows. Section 2 summarizes related studies on domain adaptation and deep learning-based engineering performance prediction methods. Section 3 presents the overall framework of the proposed model. Section 4 shows the results of the proposed model and compares the model with existing ones. Section 5 presents the conclusions and limitations of this work.

## 2. Related Works

### 2.1 Domain Adaptation in Computer Science
### 2.1.1 Knowledge Transfer

Deep learning-based methodologies with supervised learning have been proposed to solve problems, such as classification, regression, and object detection, in various fields. Predictions can be made about a new unseen problem by training a deep learning model on vast amounts of data under the assumption that the test data are drawn from the same distribution as the training data. However, collecting sufficient training data and annotating the dataset are time consuming and expensive, thus limiting the use of deep learning in industrial applications, such as 3D problems. Therefore, general methods for multiple tasks or domains that can lead to multi-task learning, multi-view learning, domain generalization, and transfer learning are needed.

In reality, the distribution of new unlabeled data often differs from that of the data used in training, known as domain shift (Quinonero-Candela et al., 2008). Even though a model with high accuracy is built based on training data, the model's prediction accuracy may decrease when the model is used for test data. Domain generalization and transfer learning are often adopted to solve this problem. They aim to train a domain-invariant model for generalization to the unseen target domain. However, they differ because transfer learning involves either sparsely labeled or unlabeled target data to align domain-level features while training. Transfer learning has been used for either similar or different tasks with the goal of providing inference about a new target domain on the basis of a model trained through the source domain.

According to Pan and Yang (2010), domain adaptation is a popular method in transductive transfer learning that aims to perform the same task in different domains, and it involves data with the same feature space but different distributions.

### 2.1.2 Domain Adaptation

A deep learning model that can solve various domains for a single task is needed to utilize deep learning in real-world problems, such as those in industrial fields, and this process is known as domain adaptation. Domain adaptation is a method of minimizing the target risk by applying knowledge transfer for test data (target domain) on the basis of the knowledge obtained from labeled data (source domain) (Huang et al., 2006; Ben-David et al., 2010). According to Farahani et al. (2021), domain adaptation is used when the feature space of different domains does not change, but their probability distributions differ. In closed-set domain adaptation, even though a domain gap exists between the source and target domains, they still share the same label space. Traditional domain adaptation usually corresponds to this condition. As Farahani et al. (2021) stated, domain shift can be categorized as prior shift, covariate shift, and concept shift. Most domain adaptation methods aim to solve covariate shift, where conditional probability distributions remain constant across domains, but the marginal probability distribution differs.

Domain adaptation can be classified into supervised, semi-supervised, and unsupervised on the basis of the presence or absence of a target domain label. Unsupervised domain adaptation applies knowledge from labeled source domain data to an unlabeled target domain. In real-world problems, unlabeled target domains are common because of the long front-loading time for collecting 3D data and annotating the labels. Therefore, in industrial fields, unsupervised domain adaptation can be used practically and reduce the data collection burden.

**2.1.3 Deep Domain Adaptation**

With the increasing usage of deep learning in approximating complex functions, domain adaptation via deep neural networks has been widely studied. This area is known as deep domain adaptation. According to Wilson and Cook (2020), domain adaptation can be classified into different types depending on how the domains are aligned: by minimizing divergence, by performing reconstruction, and by adversarial training.

The adversarial-based approach is the most popular method in deep domain adaptation. In this approach, the feature extractor is trained so that it cannot distinguish between the source and target domains by using a domain classifier, which acts as a discriminator in GAN, to extract domain-invariant features. The most representative study is that of Ganin et al. (2016), who proposed domain-adversarial neural network (DANN) to reduce H divergence through a neural network composed of a feature extractor, classifier, and domain classifier to solve classification problems domain-invariantly. By adding a gradient reversal layer in front of the domain classifier, the input is passed through in the forward pass, and a negative gradient is backpropagated so that the feature extractor cannot distinguish the domains. This method has become the basis for various adversarial-based domain adaptation methods (Tzeng et al., 2017; Laradji & Babanezhad et al., 2020).

Domain adaptation has been widely used in many fields, such as natural language processing, computer vision, and multimodal data (Singhal et al., 2023). In computer vision, it has been initially applied to simple image-handling problems, such as digit classification (MNIST, MNIST-M, SVHN, and USPS) and image classification (Office-31, Office-Home, and DomainNet). Gradually, other complex tasks, such as pose estimation (LINEMOD), gaze estimation (EYEDIAP), segmentation for medical images (Liu et al., 2021; Bateson et al., 2022), 3D object detection (Cityscapes, Foggy Cityscapes) in autonomous driving (Triess et al., 2021), and 3D point cloud segmentation (SemanticKITTI), are being attempted. Therefore, most domain adaptation studies perform classification tasks. As stated by de Mathelin et al. (2021), most deep domain adaptation studies are specialized for classification tasks, and because the regression space is continuous with no distinct boundaries unlike classification space, extending them to regression tasks is difficult.

Mansour et al. (2009) introduced a new theoretical bound for regression problems through the concept of discrepancy between predictors. On this basis, Richard et al. (2021) proposed an adversarial hypothesis-discrepancy multi-source domain adaptation (AHD-MSDA) specific to the regression task. The domain-invariant features were extracted by training the attribute weights of each source domain on the basis of the relation between each source domain and the target domain to solve multi-source domain adaptation. This is because different source domains have different relations with the target domain. In addition, this work adopted hypothesis discrepancy rather than divergence to measure the similarity between domain distributions, which is suitable for regression tasks. Similar to the domain classifier in DANN, predictor $h'$ is similar to predictor $h$ for the source domain, but it differs for the target domain. Using these networks, the hypothesis discrepancy between the weighted source and target domains is reduced. However, because they employed the same weight for source instances belonging to the same domain, they did not consider the variations inside a domain. Given that each source instance may have different relations to the target domain even though they belong to the same domain, this weighting strategy might cause negative transfer for some cases.

de Mathelin et al. (2021) also used an instance-based approach specialized to regression for supervised domain adaptation. To reduce y-discrepancy, they proposed an adversarial weighting approach that adopts weighting neural networks. The authors claimed that instance-based methods can prevent negative transfer. In addition, some studies have utilized domain adaptation to solve regression problems in pose estimation (Jiang et al., 2021) and gaze direction (Wang et al., 2022). These studies showed that the discrepancy method is more effective than the divergence method for regression tasks.

**2.2 Engineering Performance Prediction**

In the product design process of manufacturing, the engineering performance of various designs must be inspected to ensure safety. Product development requires the repetitive process of modeling prototypes, evaluating performance, and revising the design, which make the process costly and inefficient. Computer simulations, such as FEA and CFD, are utilized to replace the actual inspection test during the initial product development stage. However, computer simulations still require high computational costs and are time consuming.

**2.2.1 Deep Learning-based Prediction**

In industrial product development, deep learning-based engineering performance prediction models can replace conventional FEA, allowing for a diverse range of design options to be examined in the early conceptual design phase. Therefore, recent studies have predicted the engineering performance of structures by using deep learning methods (Liang et al., 2018; Nie et al., 2020). For example, Khadilkar et al. (2019) proposed a stress prediction model for bottom–up stereolithography 3D printing, Yoo et al. (2021) predicted the natural frequency of 2D wheel images, and Shin et al. (2023) predicted the magnitude of the maximum von Mises stress and the location of 3D road wheels.

**2.2.2 Deep Domain Adaptation for Engineering Performance Prediction**

When using deep learning-based engineering performance prediction models for industrial product development, the trained models are tested on newly developed products by assuming that new products will have the same distribution as the training data. However, because 3D design data are composed of many complex features, when a product with a slightly different design (target domain) is tested, the probability of obtaining incorrect predictions is high even with a model that has high accuracy for training data. Therefore, prediction models must be constructed through domain adaptation in industrial applications. For instance, Shin et al. (2023) proposed a prediction model composed of pretrained 3D convolutional variational autoencoder (cVAE) and a 2D convolutional autoencoder (cAE) network to predict the engineering performance of 3D road wheels and replace wheel impact analysis. Even though the proposed model has high accuracy for the concept wheels (training data), the prediction accuracy diminishes when predicting detailed wheels (unseen wheels) despite the use of transfer learning with some target domain labels. Limited domain adaptation research has been conducted on engineering problems (Liu et al., 2019; Zhang et al., 2021a; Zhang et al., 2021b), and no study has utilized domain adaptation in a deep learning-based engineering performance prediction model to replace conventional FEA.

Given that engineering problems often deal with multi-source domain data in which various source domains with different distributions exist, unsupervised domain adaptation methods that can utilize multi-source domain knowledge to perform multi-output regression tasks in an unlabeled target domain are needed. In addition, the geometry and engineering performance features of complex 3D design data must be considered in source instance weighting rather than simply using trainable weights.

In sum, domain adaptation methods for engineering analysis problems should consider the following: unlabeled target domain data, multi-source domain data, various geometry features and engineering performance of the 3D design, and the multi-output regression task.

Therefore, we propose a BW-UDA method that considers the geometry features and engineering performance of 3D design data. The proposed method is specialized for engineering analysis problems with multi-source domains. We apply the proposed method to the problem of predicting wheel impact performance. For road wheels, even if the spoke design is similar, the data distribution will be slightly different for each rim cross section. Given that the shape of the rim cross section is an important feature of a road wheel, we construct a general prediction model to predict the impact performance of wheels with different rim cross sections. For multi-source domain problems, previous studies have used multi-source domain adaptation (Mansour et al., 2008) to solve classification problems (Zhao et al., 2018; Peng et al., 2019; Liu & Ren, 2022; Wu et al., 2023) and regression problems (Richard et al., 2021). However, classifying the source domains for all new data in engineering problems is difficult, and using multiple networks as much as the amount of source domains is computationally expensive. Thus, we present a source instance weighting method called bi-weighting strategy to assign high weights to source instances with high relation to the target domain in consideration of the geometry features and engineering performance of the 3D design data. Although our method is not specialized for multi-source domain adaptation, we show that our method is powerful even for multi-source domain problems by conducting an experiment on a multi-source domain 3D wheel dataset.

# 3. Methodology

In this section, the proposed BW-UDA method is explained in detail. In Section 3.1, the 3D wheel dataset used for the experiment is described. In Section 3.2, details about the proposed method and the training process are provided.

## 3.1 3D Design Data

The task of this unsupervised domain adaptation is to predict the magnitude of the maximum von Mises stress and the location (x, y, and z coordinates) of 3D road wheels with different rim cross sections. Therefore, 3D wheel CAD with different rim cross sections and the corresponding engineering performance value obtained through a wheel impact analysis should be collected. The wheel dataset was similar to the dataset used by Shin et al. (2023). 3D concept wheel CAD was generated with the 3D CAD automation framework proposed by Yoo et al. (2021) by using 1,015 different 2D disk-view spoke design images and five types of representative rim cross sections. The 3D wheel voxel data and the impact barrier mass (kg) value used in the impact analysis were employed as the inputs, and the magnitude and location of the maximum von Mises stress were used as the labels.

For the input data, the 3D wheel dataset consists of five types of rim cross sections, as shown in Figure 1. The lower part of the rim was cropped, and only the cross section of the upper part of the rim was used because the lower part of the rim has the same shape, and it is usually not considered in the wheel impact analysis. Therefore, only the important part of the wheel, which is the wheel spoke body, was extracted, as shown in Figure 2. For each rim type, 271 wheel data for Rim 1, 332 wheel data for Rim 2, 469 wheel data for Rim 3, 467 wheel data for Rim 4, and 476 wheel data for Rim 5 were collected. A total of 2,015 3D wheel spoke CAD were collected and converted into a $64 \times 64 \times 64$ voxel.

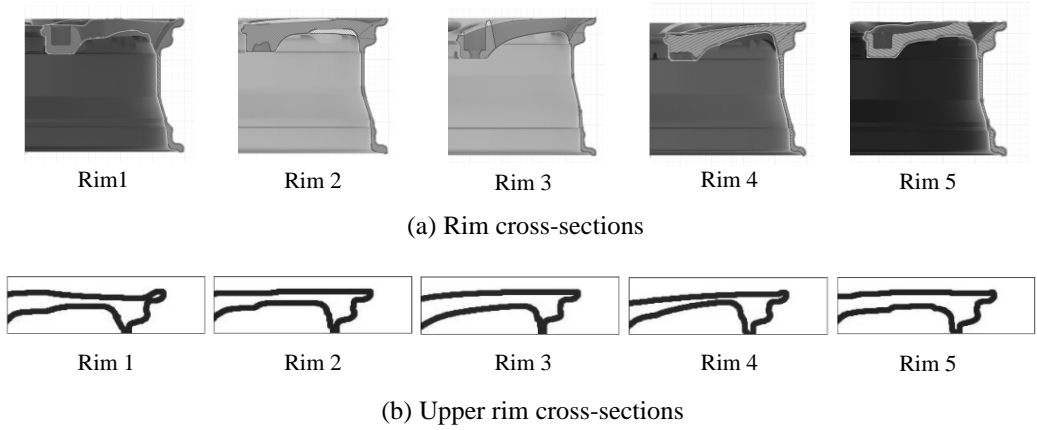

**Figure 1**   Five types of rim cross sections

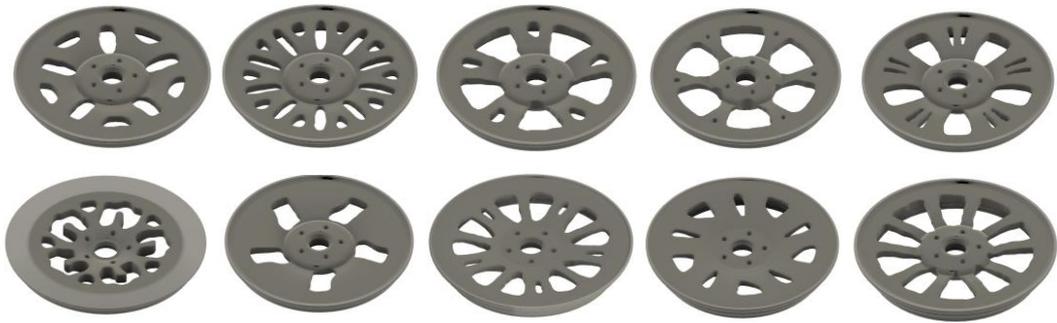

**Figure 2**   Example of a 3D wheel spoke body

The 3D wheel voxels were compressed into 2D latent space through t-distributed stochastic neighbor embedding (t-SNE; Van der Maaten & Hinton, 2008) to see the data distribution of the 3D wheel dataset, as shown in Figure 3. The figure shows that the data distribution differed for wheels with different types of rim cross sections. The type of rim cross section greatly affected the feature of a wheel. Wheels with the same rim type can be classified as a single domain because these data distribution gaps mean different domains.

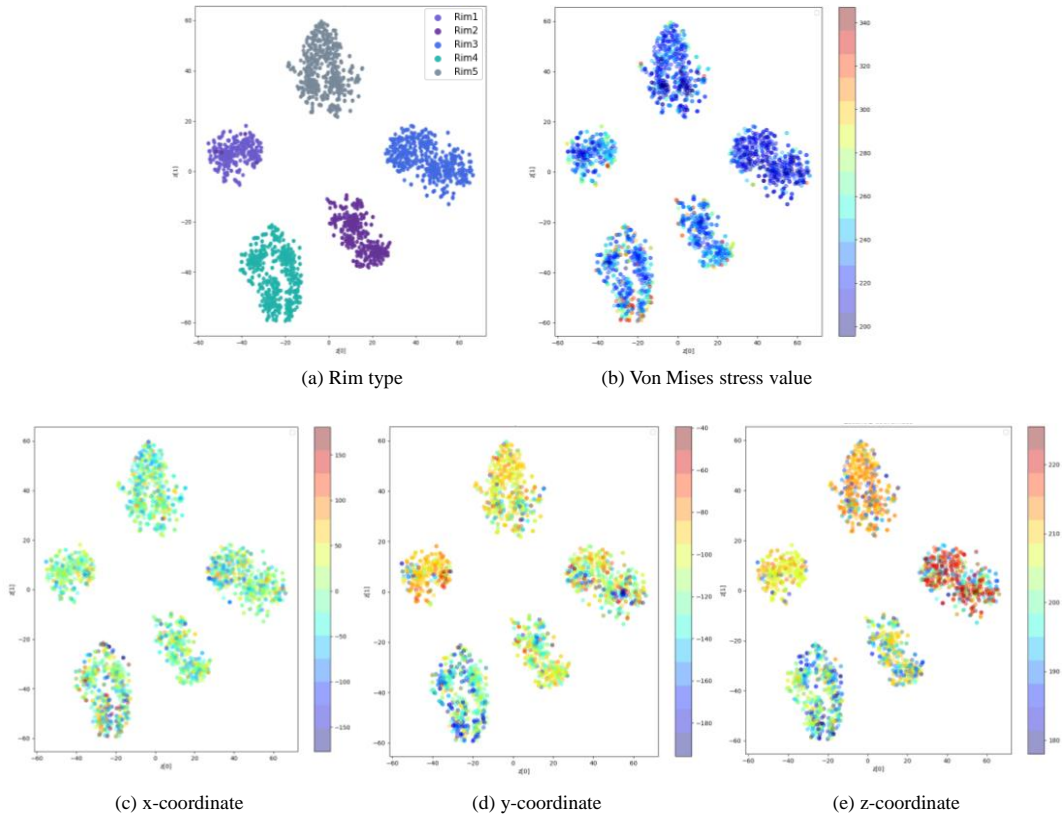

(a) Rim type     (b) Von Mises stress value

(c) x-coordinate     (d) y-coordinate     (e) z-coordinate

**Figure 3**    2D latent space of the 3D voxel wheel dataset marked by each label

A prediction model trained on a wheel with one rim cross section is useless for a concept wheel with a new rim cross section. Therefore, a sufficient amount of CAD must be collected for each rim type, the impact performance must be evaluated through FEA, then a prediction model must be trained separately for each domain to build an accurate prediction model for predicting wheel impact performance. However, collecting enough 3D CAD and label data to train a prediction model for each rim type is challenging and computationally expensive because an average time of 2 h per wheel (excluding the modeling time) is spent for FEA. The data collection and training processes must be repeated every time a wheel with a new rim cross section is created. This situation limits the use of deep learning-based engineering performance prediction models.

Therefore, we want to use domain adaptation to obtain accurate predictions for unlabeled target domains (concept wheels with new rim cross sections) on the basis of labeled source domain knowledge. This scheme utilizes existing source data and enables unsupervised learning without any label for the target domain.

The wheel data for each rim type were treated as different domains during domain adaptation. Given that our wheel dataset had five domains, we set the four domains as multi-source domains and the remaining domain as the target domain.

In addition, an appropriate range of impact barrier mass was used for the wheel impact analysis with the help of experts in the field (e.g., 498–558 kg). The distribution of the impact barrier mass for each domain is shown in Figure 4.

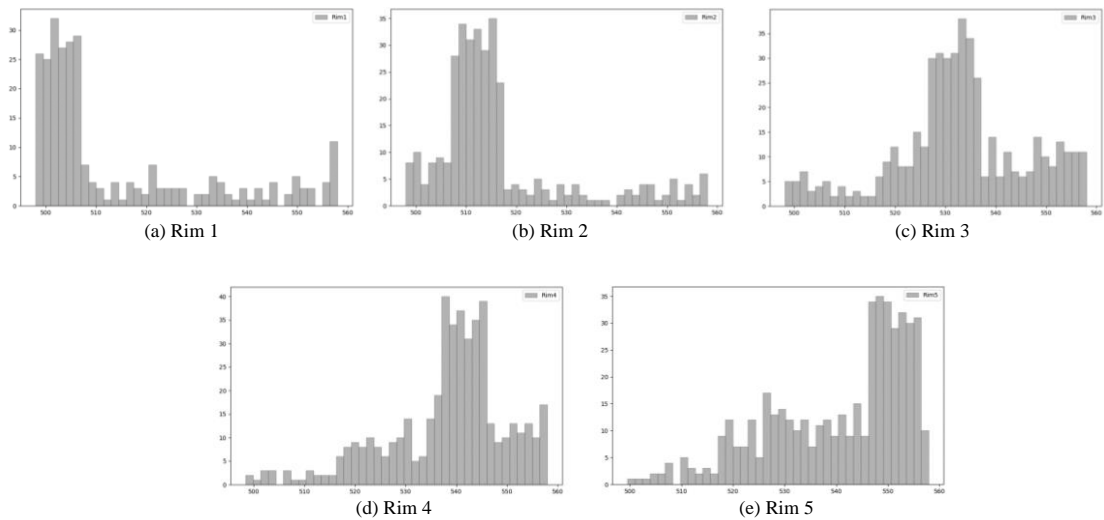

**Figure 4**  Impact barrier mass distribution of each domain

For the label data, the magnitude of the maximum von Mises stress and the x, y, and z coordinates of the corresponding location were obtained through the wheel impact analysis using the assigned impact barrier mass for each wheel. The outliers were removed from the impact analysis results. Then, the location of the maximum stress and the corresponding magnitude were extracted for each wheel data. The distribution of the label data for each domain is shown in Figure 5 (location coordinates) and Figure 6 (von Mises stress values).

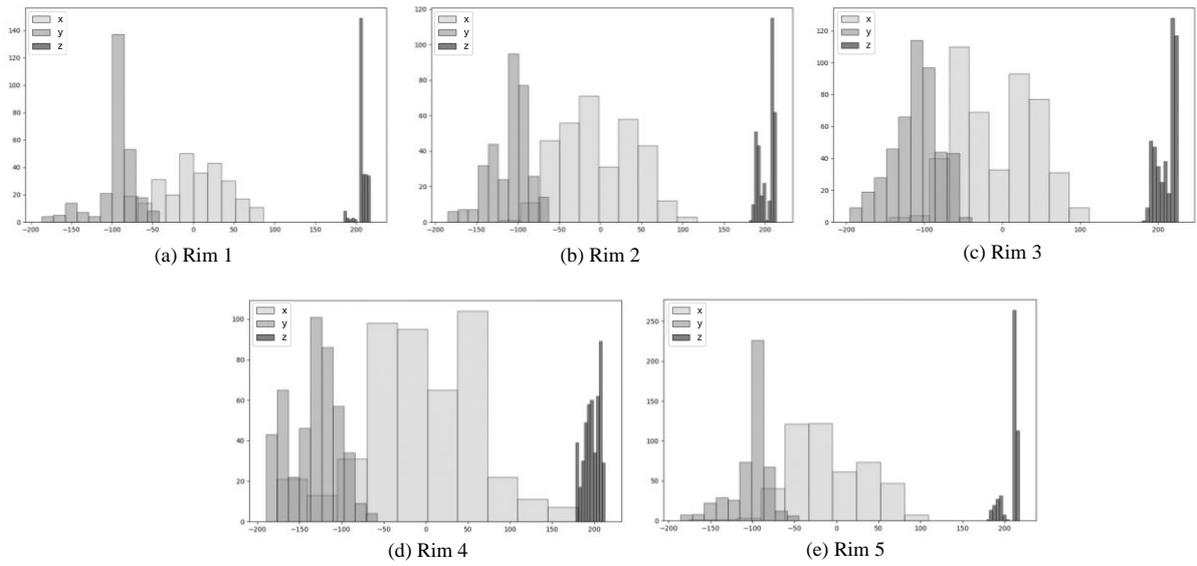

**Figure 5** Output distribution of each domain (location of maximum stress)

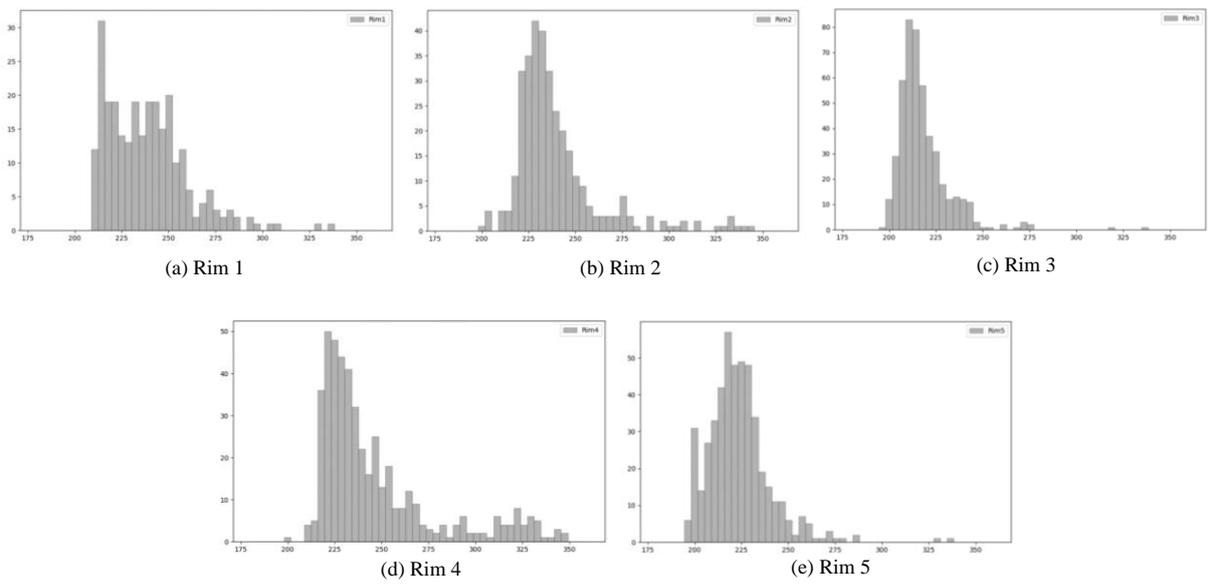

**Figure 6** Output distribution of each domain (magnitude of maximum stress)

## 3.2 BW-UDA Method

The framework of the proposed method is discussed in this section. The task of the proposed method is defined, and the model architecture is explained. Afterward, the bi-weighting strategy for the source instances and the training process are discussed.

### 3.2.1 Task

As indicated in the previous section, the 3D wheel dataset is composed of labeled multi-source domains $\{D^{s_i}\}_{i=1}^{k} = \{X_{N_i}^{s_i}, Y_{N_i}^{s_i}\}_{i=1}^{k}$ with marginal distribution $\{P_i\}_{i=1}^{k}$, where $k$ is the number of source domains, $s_i$ is the $i$th source domain, and $N_i$ is the amount of data in the corresponding source domain, and a single unlabeled target domain $D^t = \{X_{N_t}^{t}, Y_{N_t}^{t}\}$ with marginal distribution $Q$, where $N_t$ is the amount of the target domain data. Each input X contains 3D voxel $X_{voxel}$ and impact barrier mass $X_{mass}$, where $\{X_{N_i}^{s_i}\}_{i=1}^{k} = \{X_{voxel_{N_i}}^{s_i}, X_{mass_{N_i}}^{s_i}\}_{i=1}^{k}$ and $X_{N_t}^{t} = \{X_{voxel_{N_t}}^{t}, X_{mass_{N_t}}^{t}\}$. The label Y consists of the coordinate of the maximum von Mises stress occurrence and the magnitude of the stress, where $\{Y_{N_i}^{s_i}\}_{i=1}^{k} = \{Y_{x_{N_i}}^{s_i}, Y_{y_{N_i}}^{s_i}, Y_{z_{N_i}}^{s_i}, Y_{vm_{N_i}}^{s_i}\}_{i=1}^{k}$ and $Y_{N_t}^{t} = \{Y_{x_{N_t}}^{t}, Y_{y_{N_t}}^{t}, Y_{z_{N_t}}^{t}, Y_{vm_{N_t}}^{t}\}$.

The ultimate goal of this method is to minimize the target risk of the multi-output regression problem to replace the FEA for the wheel impact test. Previous studies on domain adaptation have proven that target risk can be minimized by reducing the source risk and the discrepancy between the source and target domain. Richard et al. (2021) introduced the concept of hypothesis discrepancy as a powerful method for regression problems. Hypothesis discrepancy is used to measure the similarity between distributions for any given hypothesis. Therefore, our method also uses hypothesis discrepancy to solve the regression problem with the help of the bi-weighting strategy.

Hypothesis discrepancy was used to minimize the domain gap between the source domains and the target domain and thus reduce the target risk of the regression task. An accurate regressor can be obtained even for the unlabeled target domain by minimizing the source risk and hypothesis discrepancy as in Equation 1.

$$\min_{h \in \mathcal{H}} \varepsilon_P(h, Y) + HDisc_{\mathcal{H},L}(P, Q; h) \tag{1}$$

$P$ and $Q$ are the source and target distributions over the input space, respectively; $\mathcal{H}$ is the hypothesis class with $h \in \mathcal{H}$; $\varepsilon_P(h, Y)$ is the source risk defined as $\mathbb{E}_{x \sim P}[L(h(x), Y)]$; $L$ is the loss; and $HDisc_{\mathcal{H},L}(P, Q; h)$ is the hypothesis discrepancy defined as (Richard et al., 2021)

$$HDisc_{\mathcal{H},L}(P, Q; h) = \max_{\hat{h} \in \mathcal{H}} |\mathbb{E}_{x \sim P}[L(h(x), \hat{h}(x))] - \mathbb{E}_{x \sim Q}[L(h(x), \hat{h}(x))]|. \tag{2}$$

However, because different source instances have different relations to the target domain, training the networks through Equation 1 can incur a negative transfer. A bi-weighting approach specialized for 3D design data that can transfer the domains into a meaningful common latent space will be explained in Section 3.2.3.

### 3.2.2 Model Architecture

The architecture of the proposed method consists of three networks, namely, 3D feature extractor network $\theta$, predictor network $h$, and discrepancy network $\hat{h}$, as shown in Figure 7. The overall framework of the proposed model is a multi-output regression model that predicts the magnitude of the maximum von Mises stress and the corresponding location (x, y, and z coordinates) given a 3D wheel voxel and the impact barrier mass as inputs. The model is trained to reduce the source risk while also reducing the hypothesis discrepancy between the source and target distributions to reduce the target risk.

Even though our dataset is a multi-source domain, we did not use separate predictor or discrepancy networks for each source domain like existing MSDA methods do because domain distinctions are difficult to make in engineering problems. Moreover, training a network for each of the many domains separately is too computationally expensive because additional source domains can appear in the future. Therefore, we proposed a UDA method that can utilize multi-source domains even through a single feature extractor network, a predictor network, and a discrepancy network. The method utilizes a bi-weighting strategy to weight the source instances. The weighting strategy will be described in Section 3.2.3.

Feature extractor network $\theta$ is based on a 3D convolutional neural network and extracts domain-invariant features of the 3D wheel voxel input. It consists of five 3D convolutional layers, two max-pooling layers, batch normalization layers, leaky rectified linear units (leaky ReLU) for the activation function, and a fully connected layer at the end to extract the latent features (Z) of 30 dimensions. The latent vector extracted from the feature extractor is combined with the input impact barrier mass value. Hence, a 31-dimensional intermediate input enters the predictor network and the discrepancy network.

Predictor network $h$ is a multi-output regression network that predicts the location of maximum stress (x, y, and z coordinates) and the corresponding magnitude. It consists of two fully connected layers, and a rectified linear unit (ReLU) is used for the activation function.

Discrepancy network $\hat{h}$ is trained to maximize discrepancy in the same way as the method proposed by Richard et al. (2021). The discrepancy network is trained to predict similarly as the predictor network does for the weighted source instances but differently for the target domain. Target risk can be reduced by minimizing the hypothesis discrepancy and the weighted source risk. Discrepancy network $\hat{h}$ has the same architecture as predictor network $h$.

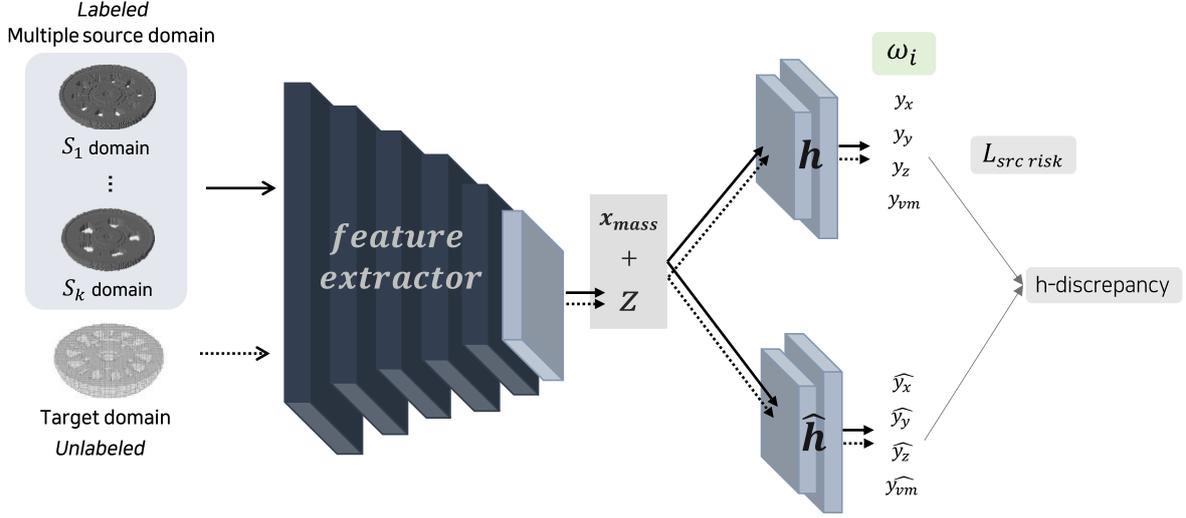

**Figure 7** Overall framework of the proposed method

The Adam optimizer was used to train the proposed model with a learning rate of 0.001, batch size of 32, and 500 epochs (with early stopping). Given that each network is trained sequentially, each is composed of individual parameters and optimizers. The training process of the three networks and the corresponding loss functions will be explained in detail in Section 3.2.4.

### 3.2.3 Bi-weighting Strategy

An appropriate weight should be assigned to each source instance because different source instances have different relations to the target domain. In particular, a method of obtaining appropriate weights in consideration of the features of each source instance is necessary. We developed a bi-weighting strategy that utilizes the attributes of 3D design data in two ways. To present a weighting method suitable for 3D engineering structures, we combined two weighting factors, namely, a weighting factor based on the geometry features $\{W_{geo\ N_i}^{s_i}\}_{i=1}^{k}$ and another weighting factor based on engineering performance $\{W_{eng\ N_i}^{s_i}\}_{i=1}^{k}$.

#### *Geometry feature-based weighting*

First, a geometry feature-based weighting strategy, $\{W_{geo\ N_i}^{s_i}\}_{i=1}^{k}$, was developed to reflect the geometry information of the 3D design data. Among the source domain instances, instances with geometry features similar to those of the target domain received a high weight. The feature distance was measured in a low-dimensional latent space to calculate the distance between the geometry features of the instances. The wheel voxels of the source domains and the target domain were compressed into a 2D space by using t-SNE to obtain the latent code of each 3D wheel sample. Next, the Euclidean distance between the mean of the target latent codes and each latent code of the source instances was calculated. The reciprocal of the obtained distance per source sample was obtained so that a source instance close to the target domain can receive a high weight. Min–max scaling was applied to normalize the weights between 0 and 1.

Geometry weighting factor was used to obtain the loss of the weighted source risk and the weighted source discrepancy in each epoch. The geometry features do not change, so the geometry weighting factor was fixed throughout the training process. Later on, this factor will be combined with the weighting factor based on engineering performance features in the training process. The loss functions will be discussed in Section 3.2.4 in detail.

### *Engineering performance feature-based weighting*

A weighting strategy was developed to consider the engineering performance of the 3D design data. This factor, $\{W_{eng\ N_i}^{s_i}\}_{i=1}^{k}$, was based on the relation of the engineering performance features of the target domain and each source instance. A source instance whose engineering performance was similar to the target domain received a high weight. For this purpose, we utilized the discrepancy between predictor $h$ and discrepancy network $\hat{h}$. Source instances with high discrepancy between the networks have similar engineering performance attributes as the target domain because discrepancy network $\hat{h}$ is trained to predict similarly as predictor $h$ for the source domains but differently for the target domain. Therefore, source instances with high discrepancy are assigned high weights. The discrepancy between predictor $h$ and discrepancy network $\hat{h}$ for each source sample was measured, and min–max scaling was applied to normalize the weights. The discrepancy can be used to assume similarity regarding the engineering performance features because the latent features that go through predictor $h$ and discrepancy network $\hat{h}$ compress the features of engineering performance (impact performance in this case). As the training progresses, the prediction accuracies of the predictor and discrepancy network increase. Therefore, the engineering performance weights are updated at the end of every epoch.

### *Combined weight*

The weighted sum of $\{W_{eng\ N_i}^{s_i}\}_{i=1}^{k}$ and $\{W_{geo\ N_i}^{s_i}\}_{i=1}^{k}$ was used for the total weight of the individual source instances. Combined weight $\omega$ was defined as $\{\alpha \cdot W_{eng\ N_i}^{s_i} + \beta \cdot W_{geo\ N_i}^{s_i}\}_{i=1}^{k}$, and appropriate coefficients α and β were adopted in an experiment. The combined weights were used to evaluate the loss of the weighted source risk and weighted source discrepancy in every epoch.

The final task of the proposed method is defined as Equation 3, where $P_\omega$ is the marginal distribution of the weighted source domains and $\varepsilon_{P_\omega}(h,Y)$ is the weighted source risk defined as $\mathbb{E}_{x \sim P_\omega}[L_\omega(h(x),Y)]$.

$$\min_{h \in \mathcal{H}} \varepsilon_{P_\omega}(h,Y) + HDisc_{\mathcal{H},L}(P_\omega, Q; h) \tag{3}$$

### 3.2.4 Training Process

The loss functions and training process of the proposed model are explained in this section. Three loss functions were used to train the model. For loss $L$, the mean squared error (MSE) was used to evaluate the error for each output. Weighted MSE is defined in Equation 4, where $n$ is the amount of data.

$$L_\omega(Y,\hat{Y}) = \frac{1}{n} \sum_{i=1}^{n} \left\{ \omega_i \cdot \frac{1}{4} \left( \left(Y_{x_i} - \hat{Y}_{x_i}\right)^2 + \left(Y_{y_i} - \hat{Y}_{y_i}\right)^2 + \left(Y_{z_i} - \hat{Y}_{z_i}\right)^2 + \left(Y_{vm_i} - \hat{Y}_{vm_i}\right)^2 \right) \right\} \tag{4}$$

$L_{src\ risk}$ was used to minimize the weighted source risk, namely, the weighted MSE, of predictor network $h$ (Equation 5).

$$L_{src\ risk} = \varepsilon_{P_\omega}(h,Y) \tag{5}$$

$L_{\hat{h}\ disc}$ was employed to maximize the hypothesis discrepancy of discrepancy network $\hat{h}$, as shown in Equation 6.

$$L_{\hat{h}\ disc} = -HDisc_{\mathcal{H},L}(P_\omega, Q; h) \tag{6}$$

$L_{feat\ disc}$ was used to minimize the hypothesis discrepancy and weighted source risk of feature extractor $\theta$, as indicated in Equation 7.

$$L_{feat\ disc} = \varepsilon_{P_\omega}(h,Y) + HDisc_{\mathcal{H},L}(P_\omega, Q; h) \tag{7}$$

The three networks were optimized sequentially in the order of the predictor network, the discrepancy network, and the feature extractor for every epoch. The training steps were as follows. First, the parameter of the predictor network was updated to minimize $L_{src\ risk}$. Second, the parameter of the discrepancy network was updated to maximize the hypothesis discrepancy by using $L_{\hat{h}\ disc}$. Third, the parameter of the feature extractor was updated to minimize $L_{feat\ disc}$. Last, the engineering performance weights of the source instances were updated using the discrepancy of the trained predictor and discrepancy networks at the end of every epoch. After training, the trained feature extractor can extract domain-invariant features of the target domain. Then, the trained predictor can predict the magnitude of the maximum von Mises stress and the location, with the extracted features and impact barrier mass as the input.

The objective of the proposed method is a min–max problem, as follows:

$$\min_{\theta,\ h \in \mathcal{H}} \max_{\hat{h} \in \mathcal{H}} \left[ \varepsilon_{P_\omega}(h, Y) + \left| \varepsilon_{P_\omega}(h, \hat{h}) - \varepsilon_Q(h, \hat{h}) \right| \right]. \tag{8}$$

# 4. Results and Discussion

The prediction results of the proposed model on the 3D wheel dataset are discussed in this section. In Section 4.1, various metrics are defined to evaluate the performance of the proposed model, and the results are visualized. In Section 4.2, the prediction results of the proposed model are compared with those of other baseline models to show the effectiveness of the proposed model.

## 4.1 Results of the Proposed Model

The proposed method showed powerful results for the 3D wheel dataset through a compact network even though the wheel dataset contained multi-source domain data. Various metrics were used to evaluate the performance of the proposed model. Root mean square error (RMSE) and mean absolute percentage error (MAPE) were adopted to evaluate the prediction accuracy of the maximum von Mises stress value, and they are defined in Equations 9 and 10, respectively. $Y_{vm_i}$ is the ground truth maximum stress value, $\hat{Y}_{vm_i}$ is the prediction maximum stress value, and $n$ is the total number of the dataset.

$$RMSE_{von\,Mises} = \sqrt{\frac{1}{n}\sum_{i=1}^{n}(Y_{vm_i} - \hat{Y}_{vm_i})^2} \qquad (9)$$

$$MAPE_{von\,Mises} = 100 \times \frac{1}{n}\sum_{i=1}^{n}\left|\frac{Y_{vm_i} - \hat{Y}_{vm_i}}{Y_{vm_i}}\right| \qquad (10)$$

Euclidean distance error was used to evaluate the prediction of the maximum stress location. The means and medians of the 3D Euclidean distance error and 1D Euclidean distance error were evaluated to compare the prediction results for each x, y, and z coordinate in detail. The equation for the 3D Euclidean distance error is shown in Equation 11, where $x_i$, $y_i$, and $z_i$ are the ground truths for x, y, and z coordinates, respectively, and $\hat{x}_i$, $\hat{y}_i$, and $\hat{z}_i$ are the predictions for x, y, and z coordinates, respectively.

$$Euclidean\ Distance_{3D} = \sqrt{(x_i - \hat{x}_i)^2 + (y_i - \hat{y}_i)^2 + (z_i - \hat{z}_i)^2} \qquad (11)$$

### *Prediction Results*

In the experiment, each of the five domains was set as the target domain. The results of the proposed model are organized in Table 1. The average results of the proposed model show that the mean 3D Euclidean distance error was 53.620 mm, and the median 3D Euclidean distance error was 49.399 mm. The prediction and ground truth maximum stress locations are visualized in Figure 8. The figure shows that most of the maximum von Mises stress occurred at the crack near the spoke hole. The MAPE of the maximum von Mises stress value was 4.701%, and its RMSE was 14.6 MPa.

According to the results of each target domain, the prediction accuracy was the lowest when Rim 4 was set as the target domain. This outcome can be attributed to the fact that the distribution of x, y, and z coordinate labels in this domain data differed from that in the other domain data, as shown in Figure 5. In addition, Rim 4 had the thinnest rim cross section and the largest prediction error for maximum stress values.

**Table 1**  Proposed model prediction results

| Target Domain | 3D Euclidean Distance Error (Mean) (mm) | 3D Euclidean Distance Error (Median) (mm) | Von Mises MAPE (%) | Von Mises RMSE (MPa) |
|---|---|---|---|---|
| Rim 1 | 44.991 | 40.319 | 4.746 | 14.959 |
| Rim 2 | 50.192 | 47.136 | 4.551 | 14.777 |
| Rim 3 | 54.854 | 54.846 | 3.842 | 11.412 |
| Rim 4 | 66.283 | 57.925 | 6.864 | 21.183 |
| Rim 5 | 51.779 | 46.769 | 3.501 | 10.669 |
| **Avg** | **53.620** | **49.399** | **4.701** | **14.600** |

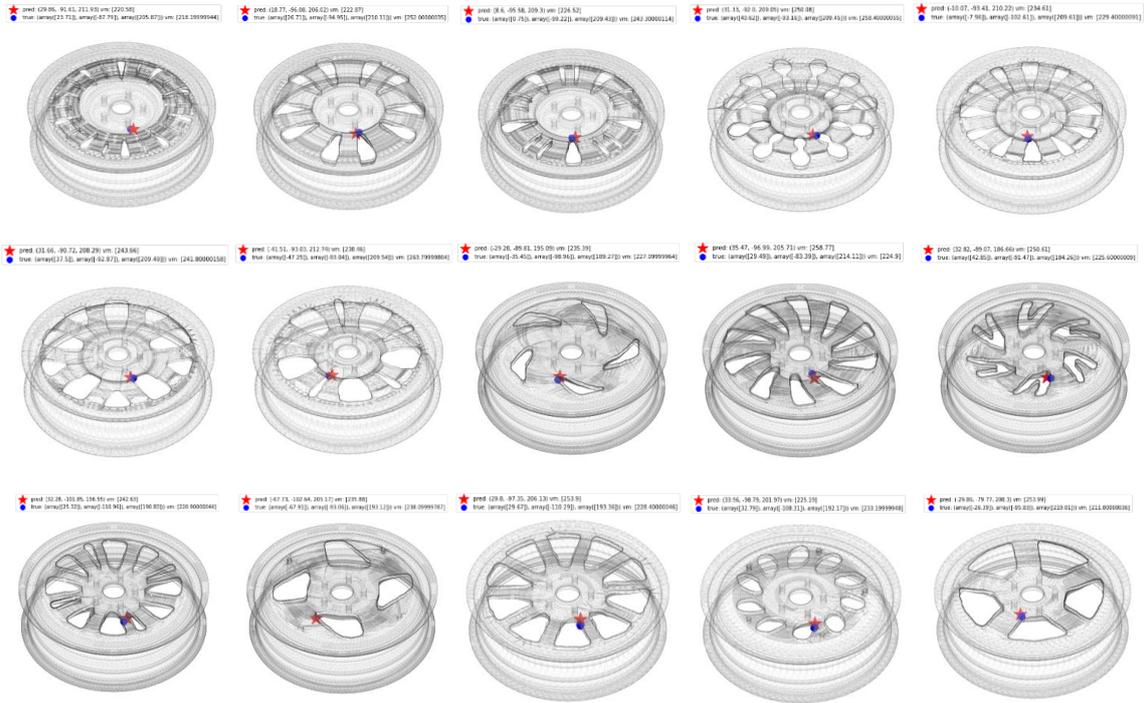

**Figure 8**    Visualization of the prediction location (red star) and ground truth location (blue dot)

Given that the prediction error for the location of the maximum stress was relatively large, we further analyzed the location predictions through the 1D Euclidean distance error of the x, y, and z coordinates individually. The results of the 1D Euclidean distance error for each target domain are shown in Table 2. The average results of the proposed model indicate that the mean 1D Euclidean distance errors were 44.521, 20.173, and 8.204 mm, and the median 1D Euclidean distance errors were 40.458, 15.110, and 7.042 mm for the x, y, and z coordinates, respectively. The prediction for the z coordinate was the most accurate because the z axis was the direction of the thickness of the wheel rim cross section, resulting in an easy prediction. By contrast, the prediction error was the largest for the x coordinate, which was the most challenging coordinate to predict, because the wheel was symmetrical along the x axis relative to the impact location.

**Table 2** Proposed model's 1D coordinate prediction results

| Target Domain | X 1D Distance Error (Mean) | X 1D Distance Error (Median) | Y 1D Distance Error (Mean) | Y 1D Distance Error (Median) | Z 1D Distance Error (Mean) | Z 1D Distance Error (Median) |
|---|---|---|---|---|---|---|
| Rim 1 | 38.285 | 34.175 | 16.268 | 11.138 | 5.114 | 4.204 |
| Rim 2 | 41.518 | 39.124 | 18.133 | 13.858 | 8.763 | 6.908 |
| Rim 3 | 44.084 | 43.862 | 22.892 | 17.651 | 10.549 | 10.259 |
| Rim 4 | 56.351 | 46.969 | 23.859 | 18.875 | 8.358 | 7.064 |
| Rim 5 | 42.366 | 38.158 | 19.713 | 14.030 | 8.235 | 6.776 |
| **Avg** | **44.521** | **40.458** | **20.173** | **15.110** | **8.204** | **7.042** |

We assume that the proposed model is well trained, as intended by domain adaptation. In this case, the trained 3D feature extractor should be able to extract the latent space where the domain-invariant features are aligned by their engineering performance level. The latent space of the source and target domains extracted through the trained feature extractor is shown in Figure 9 (with Rim type 3 as the target domain). As presented in Figure 3, the original distributions of the initial wheel dataset were clustered in accordance with each rim type. As a result, gaps existed between domains, and each source domain distribution was not aligned based on the engineering performance features. However, in the latent space extracted by the trained model, the source and target domains were meaningfully transferred to one common space. In particular, in Figure 9(f), they are aligned in accordance with the magnitude of the maximum von Mises stress. Wheel data with large maximum von Mises stress values are marked in red, and those with small maximum von Mises stress values are marked in blue. Wheel data with high von Mises stress values are aligned toward the right part of the figure.

The latent space for each target domain case is shown in Figure 10. Given that the latent spaces are particularly aligned by their maximum stress values, the plots are colored by their domains (rim types) and maximum stress values. This figure may explain why the prediction accuracy for von Mises stress improved dramatically.

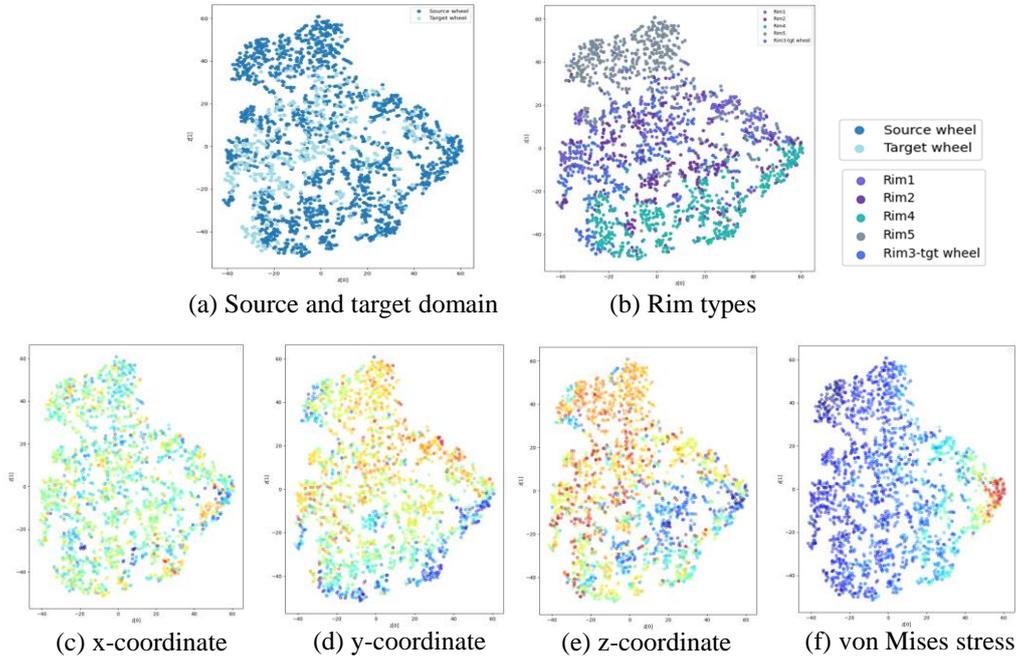

(a) Source and target domain  (b) Rim types

(c) x-coordinate  (d) y-coordinate  (e) z-coordinate  (f) von Mises stress

**Figure 9**　Latent space of the dataset marked by each label

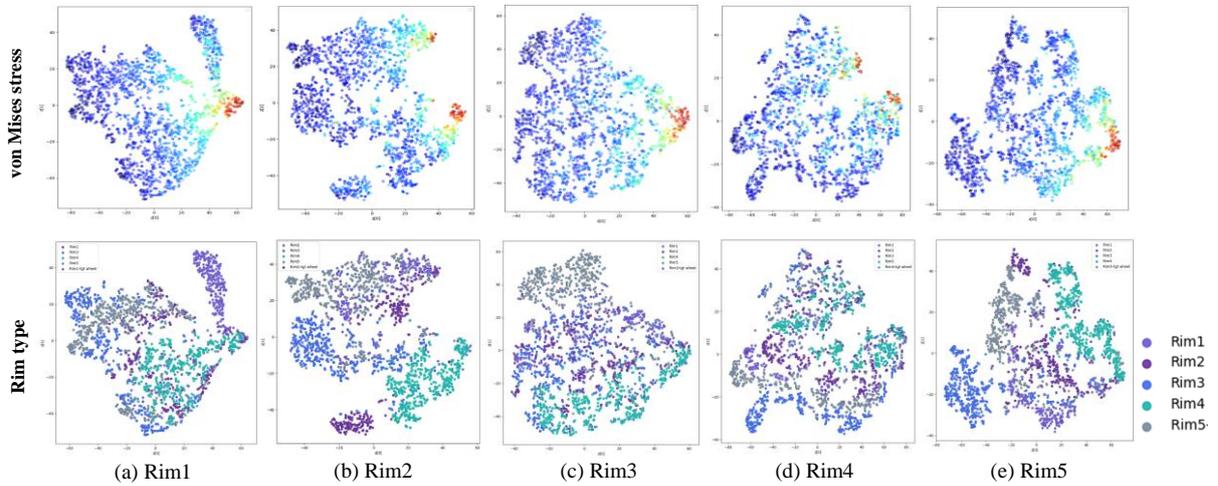

(a) Rim1  (b) Rim2  (c) Rim3  (d) Rim4  (e) Rim5

**Figure 10**　Latent space of the dataset for each target domain case

## *Comparison of weighting factors*

The results of the proposed model without any weights and with only one of each weight were compared to understand the importance of the proposed bi-weighting strategy to this method. The model architecture and training conditions were similar to those in the proposed model, except for the weighting factors. The results are shown in Table 3.

Table 3   Prediction results obtained by varying the weights

| Target Domain (Avg) | 3D Euclidean Distance Error (Mean) (mm) | 3D Euclidean Distance Error (Median) (mm) | Von Mises MAPE (%) | Von Mises RMSE (MPa) |
|---|---|---|---|---|
| No weights | 55.899 | 51.566 | 7.582 | 24.021 |
| Only $W_{geo}$ | 55.524 | 50.963 | 6.308 | 22.261 |
| Only $W_{eng}$ | 54.628 | 50.079 | 6.264 | 19.975 |
| Both weights | **53.620** | **49.399** | **4.701** | **14.600** |

The average prediction results of the model with no weights show that the mean 3D Euclidean distance error was 55.899 mm, the median 3D Euclidean distance error was 51.566 mm, the MAPE of the maximum stress value was 7.582%, and the RMSE of the maximum stress value was 24.021 MPa. For the model with only the geometry feature-based weight, the mean 3D Euclidean distance error was 55.524 mm, the median 3D Euclidean distance error was 50.963 mm, the MAPE of the maximum stress value was 6.308%, and the RMSE of the maximum stress value was 22.261 MPa. The average prediction results of the model with only the engineering performance-based weight had high accuracy, where the mean 3D Euclidean distance error was 54.628 mm, the median 3D Euclidean distance error was 50.079 mm, the MAPE of the maximum stress value was 6.264%, and the RMSE of the maximum stress value was 19.975 MPa. These results show that the engineering performance feature-based weight could assign meaningful weights to the important source instances to minimize the target risk. However, the weighted combination of two weighting factors can be more effective than using only one weighting factor, especially in predicting the magnitude of the maximum von Mises stress. Therefore, the proposed bi-weighting strategy is a powerful method of predicting engineering performance.

To demonstrate the effectiveness of the proposed method, we compared the prediction results of the proposed model with a basic regression model without domain adaptation, DANN (Ganin et al., 2016), and AHD-MSDA (Richard et al., 2021). The comparison results are discussed in the next section.

**4.2 Comparison with Baseline Models**

As mentioned in Section 4.1, the proposed model has high prediction accuracy for the target domain. To determine how powerful the proposed model is, we adopted the same training conditions and compared the prediction results of a basic regression model without domain adaptation, DANN (Ganin et al., 2016) that reduces H-divergence, and AHD-MSDA (Richard et al., 2021) that reduces hypothesis-discrepancy but with trainable weights for each source domain. The basic regression model was composed of only the feature extractor and the predictor network with MSE as the loss function. After training the regression model by using multi-source domain data, the prediction results were tested for the target domain. For DANN, we changed the label predictor to a regression network and used MSE for the loss function. The comparison of the prediction results is shown in Table 4.

Table 4    Comparison of baseline model prediction results (average results of the target domains)

| Methods | 3D Euclidean Distance Error (Mean) (mm) | 3D Euclidean Distance Error (Median) (mm) | Von Mises MAPE (%) | Von Mises RMSE (MPa) |
|---|---|---|---|---|
| Regression w/o DA | 60.748 | 56.182 | 8.180 | 24.472 |
| DANN | 60.727 (-0.03%) | 54.953 (-2.24%) | 8.443 (+3.12%) | 26.671 (+8.24%) |
| AHD-MSDA | 57.006 (-6.56%) | 53.084 (-5.84%) | 6.418 (-27.45%) | 21.231 (-15.27%) |
| **Proposed** | **53.620 (-13.29%)** | **49.399 (-13.73%)** | **4.701 (-74.01%)** | **14.600 (-67.62%)** |

The average prediction results of the basic regression show that the mean 3D Euclidean distance error was 60.748 mm, the median 3D Euclidean distance error was 56.182 mm, the MAPE of the maximum stress value was 8.18%, and the RMSE of the maximum stress value was 24.472 MPa. Similarly, when DANN was used, the mean 3D Euclidean distance error was 60.727 mm, the median 3D Euclidean distance error was 54.953 mm, the MAPE of the maximum stress value was 8.443%, and the RMSE of the maximum stress value was 26.671 MPa. Given that DANN is a well-suited methodology for classification problems, even with domain adaptation, the prediction results were not considerably different from the basic regression prediction results. As shown in the latent space for each target case of DANN in Figure 11, negative transfers occurred, and the latent space was not meaningfully aligned with regard to engineering performance. This situation decreased the prediction accuracy.

In addition, the average prediction results of AHD-MSDA show that the mean 3D Euclidean distance error was 57.006 mm, the median 3D Euclidean distance error was 53.084 mm, the MAPE of the maximum stress value was 6.418%, and the RMSE of the maximum stress value was 21.231 MPa. Although the prediction accuracy of AHD-MSDA improved relative to that of the basic regression model, the proposed model had the highest prediction accuracy for both the location and magnitude of the maximum von Mises stress. Thus, the proposed bi-weighting strategy based on the geometry and engineering performance features of source instances is the appropriate method, rather than the method of simply training weights for each source domain.

The MAPE of the maximum stress value was reduced by 74.01%, and the 3D Euclidean distance error for the location of the maximum stress decreased by about 7 mm compared with those in the basic regression model. This finding proves that the proposed method is an unsupervised domain adaptation method specialized for engineering analysis problems.

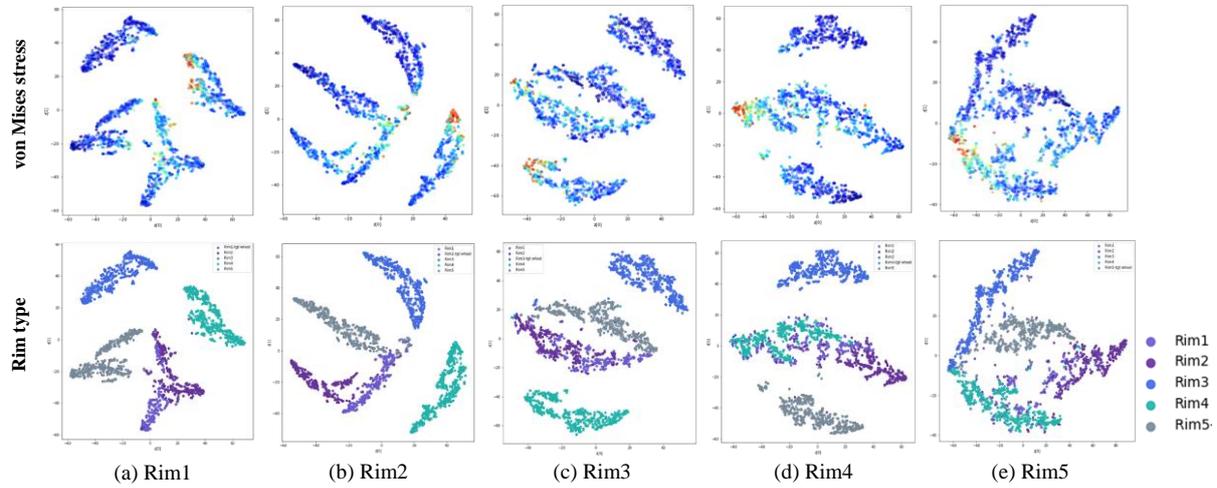

**Figure 11**   Latent space extracted through DANN

The abovementioned results confirm that the proposed model has high prediction accuracy for the unseen target domain. The training time of the proposed model is about 130 min on average (NVIDIA A100 GPU), and the trained model consumes less than a second to infer one data sample. We proved that the proposed model can be much more effective than conventional wheel impact analysis, which consumes an average of 2 h or more (64-core CPU) per wheel, while maintaining high accuracy for the unseen target domain.

# 5. Conclusion

Deep learning-based engineering performance prediction methodologies have remarkable advantages because they can dramatically reduce the time and cost of product design optimization. Thus, various studies have attempted to use these methodologies as a replacement for the conventional analysis process efficiently. However, industrial problems are constantly generating diverse 3D design data, so even though a prediction model is trained to be accurate on training data, its prediction performance degrades in domains with similar but different data distributions.

The proposed model is an unsupervised domain adaptation method specialized for engineering analysis problems. It assigns appropriate weights to each multi-source domain instance through a bi-weighting strategy. We used a method that assigns high weights to source samples with high similarity to the target domain on the basis of the geometry features and engineering performance of the 3D design data. The weighted unsupervised domain adaptation method can reduce the target risk even when the target domain has no labels by reducing the source risk and domain gap between the source and target domains and by focusing on source instances similar to the target domain. The proposed method can be applied to any engineering analysis problem.

We constructed a regression model to predict wheel impact performance on the basis of 3D wheel dataset that can replace conventional wheel impact analysis. The model can predict the location and magnitude of the maximum von Mises stress by using the 3D wheel voxel and impact barrier mass employed in the impact analysis as the input. A model consisting of a 3D feature extractor network, a predictor network, and a discrepancy network was built to reduce the distribution gap between the source and target domains by using hypothesis discrepancy, which is suitable for regression tasks.

On the basis of multi-source 3D wheel data, the prediction results of the proposed model were compared with a basic regression model, DANN and AHD-MSDA. Among the compared models, the proposed model showed the most accurate predictions for the target domain. In addition, the analysis of the weights used in the training process confirmed that the presented bi-weighting strategy that considers the weights of geometry- and engineering performance-based features had the highest prediction accuracy.

The contributions of the proposed method are also summarized here. First, to the best of our knowledge, this study is the first to apply domain adaptation to the problem of deep learning-based engineering performance prediction. Based on the knowledge of the existing source domain, unsupervised domain adaptation enables accurate engineering performance predictions even for an unlabeled target domain. This method can overcome the limitations of 3D deep learning in engineering, namely, lack of 3D data and burdensome data collection, and can therefore increase the applicability of deep learning in engineering analysis fields. In particular, the method is highly scalable, so it can be utilized in any engineering analysis problem.

Second, we developed a bi-weighting scheme specialized for 3D design data. The training process was implemented based on meaningful weights at the engineering level by using a source instance weighting scheme that reflects the geometry features and engineering performance of 3D design data. This approach is a more appropriate weighting scheme for engineering analysis problems compared with simple weighting using trainable parameters.

Last, the proposed model can be applied to multi-source domain data. Multi-source domain data are abundant, especially in industrial problems. Therefore, utilizing multi-source domain knowledge is advantageous and can solve the problem of data scarcity. The network is compact because source instance weighting is adopted, and the training proceeds in accordance with the importance of each source sample to the target domain to avoid negative transfer.

Despite these advantages, our method still has limitations. Given that the current wheel dataset was generated from five rim cross sections, the trained model was experimented on by using five domains. To generalize the proposed model, we plan to extend it to other diverse design factors, such as design parameters, to utilize domain adaptation for 3D design data. In addition, we wish to apply the proposed method to other engineering analysis problems to investigate its scalability.

**Acknowledgements**

This work was supported by the National Research Foundation of Korea grant (2018R1A5A7025409) and the Ministry of Science and ICT of Korea grant (No.2022-0-00969, No.2022-0-00986)